\pdfoutput=1
\documentclass[11pt]{article}

\usepackage{EMNLP2022}

\usepackage{times}
\usepackage{latexsym}

\usepackage[T1]{fontenc}
\usepackage[utf8]{inputenc}

\usepackage{microtype}

\usepackage{inconsolata}

\usepackage{siunitx}
\usepackage{graphicx}
\usepackage{array}
\usepackage{float}

\usepackage{booktabs}
\usepackage{etoolbox}
\renewcommand{\bfseries}{\fontseries{b}\selectfont}
\newrobustcmd{\B}{\bfseries}
\usepackage{enumitem}

\newcolumntype{H}{>{\setbox0=\hbox\bgroup}c<{\egroup}@{}}

\usepackage{listings}
\def\byt5{By\textsc{T5}}
\def\mt5{m\textsc{T5}}
\def\taf{\textsc{TaF}}

\def\massive{\textsc{MASSIVE}}

\title{Evaluating Byte and Wordpiece Level Models\\for Massively Multilingual Semantic Parsing}

\author{Massimo Nicosia and Francesco Piccinno \\
Google Research, Z\"urich\\
\texttt{\{massimon,piccinno\}@google.com}
}

\begin{document}
\maketitle
\begin{abstract}

Token free approaches have been successfully applied to a series of word and span level tasks. In this work, we compare a byte-level (\byt5) and a wordpiece based (\mt5) sequence to sequence model on the 51 languages of the \massive{} multilingual semantic parsing dataset. We examine multiple experimental settings: (i) zero-shot, (ii) full gold data and (iii) zero-shot with synthetic data. By leveraging a state-of-the-art label projection method for machine translated examples, we are able to reduce the gap in exact match accuracy to only $5$ points with respect to a model trained on gold data from all the languages.
We additionally provide insights on the cross-lingual transfer of \byt5{} and show how the model compares with respect to \mt5{} across all parameter sizes. 

\end{abstract}

\section{Introduction}

Semantic parsers map natural languages utterances into logical forms (LFs). In the context of conversational agents \cite{artzi2011bootstrapping}, robotics \cite{dukes2014semeval} or question answering systems \cite{berant2013semantic}, task-oriented semantic parsers map user queries (e.g. \emph{``set an 8 am alarm''}) to machine readable LFs (e.g. \texttt{[IN:CREATE\_ALARM [SL:TIME 8 am ]]}), in the form of structured interpretations that can be understood and executed by downstream components.
Learning parsers requires training data in the form of <utterance, LF> pairs. Such data is costly to obtain especially at large scale \cite{berant2013semantic}, since expert annotators have to derive the correct LFs given an input utterance. This problem is exacerbated in a multilingual setting, where the availability of annotators, especially for non top-tier languages, is scarce and therefore even more expensive.

With the release of \massive{}~\cite{massive}, the research community has now access to a massively multilingual semantic parsing dataset that can be used to evaluate large language models fine-tuned on the task and to study cross-lingual transfer for numerous languages.

On the multilinguality front, token-free models with byte or character based vocabularies have gained strength given their competitiveness with respect to traditional subword-based pretrained language models. Models such as \byt5{}~\cite{xu-etal-2020-end}, Canine \cite{clark-etal-2022-canine} and the Charformer~\cite{tay2022charformer} have been applied to popular multilingual benchmarks obtaining state-of-the-art results.

In this paper, we perform the first in-depth evaluation of a token-free model in the context of multilingual semantic parsing. We compare the \byt5{} and \mt5{}~\cite{xue2021mt5} models across different parameter sizes and data regime settings. In addition to that, we build a map of the cross-lingual transfer for all the languages in \massive{}.
Lastly, we show that with the use of machine translated synthetic data the accuracy of a state-of-the-art multilingual parser can be just $5$ points lower than the same parser trained with all the available multilingual supervision. To incentivize research on synthetic data augmentation approaches, we release the \massive{} English training utterances translated to 50 languages.\footnote{We release the translations in 50 languages of the \massive{} English training examples obtained with an in-house translation system at \url{https://goo.gle/massive-translations}}

\section{The \massive{} Dataset}

\massive{} \cite{massive} is a semantic parsing dataset covering $51$ languages, $18$ domains, $60$ intents and $55$ slots. The dataset was created by professional translators starting from the English SLURP dataset \cite{bastianelli-etal-2020-slurp}. A significant portion of the translations have been localized too, following the recent trend in multilingual benchmarks of replacing western-centric entities with entities that are more relevant for the target languages \cite{lin2021bitod, ding-etal-2022-globalwoz, https://doi.org/10.48550/arxiv.2201.13405}.

\subsection{Pre and Post Processing}

The annotated instances in the \massive{} dataset come in the following format:

\begin{lstlisting}
intent: alarm_set
annot_utt: despiértame a las [time : nueve de la mañana] el [date : viernes]
\end{lstlisting}

To shorten the target output and save the model from generating and potentially hallucinating unnecessary words, we map the former to the following format taken from MTOP \cite{li-etal-2021-mtop}:

\begin{lstlisting}
[IN:ALARM_SET [SL:TIME nueve de la mañana ] [SL:DATE viernes ] ]
\end{lstlisting}

For evaluation, we use a simple inverse post-processing step based on string matching to convert the model outputs back to \massive{} format.

\subsection{Synthetic Data with Translate-and-Fill}

A common approach to create multilingual synthetic data from available examples is to use machine translation \cite{moradshahi-etal-2020-localizing, sherborne-etal-2020-bootstrapping}. Utterances are translated and LF annotations are projected using word aligners and noise reduction heuristics. We instead adopt the approach from \citet{nicosia-etal-2021-translate-fill}, Translate-and-Fill (\taf{}), a label projection method in which a filler model reconstructs the full LF starting from an utterance and its LF signature.

We train an \mt5{}-xxl filler model on English instances and then directly generate the LFs of translated examples in a zero-shot fashion. Since the slot order between English and translated utterances may differ, we canonicalize the generated synthetic interpretations reordering the slots as they would occur in the translations. We have also noticed in the filler output that for some languages the slot boundaries may fall inside words. For languages with white space tokenization, we move slot boundaries to word boundaries if needed.

As an example, given an input utterance \emph{``despiértame a las nueve el viernes''} and \texttt{[IN:ALARM\_SET [SL:DATE el vier ] [SL:TIME nueve ] ]} as LF, the process looks as follows. First the arguments are reordered according to the order of appearance in the original sentence: \texttt{[IN:ALARM\_SET [SL:TIME nueve ] [SL:DATE vier ] ]}. Then slot boundaries that fall within words are extended, correcting the prediction for the second argument from \texttt{[SL:DATE vier ]} to \texttt{[SL:DATE viernes ]}. 

\section{Experiments}
\label{sec:experiments}

We use \massive{} as a test bed for two model families, \byt5{} and \mt5{}, evaluating them at all sizes in three different data settings. We report \emph{Intent Accuracy} (IA) and \emph{Exact Match} (EM) accuracy. We do not perform any hyper-parameter tuning: we train for \si{30K} steps with a fixed learning rate of 0.0001 and a batch size of $128$ for all models but xxl, for which batch size was reduced to $32$. We run fine tuning on Cloud TPU v3 with an input/target length of $1024$/$512$ for \byt5{} and $512$/$512$ for \mt5{}. To minimize compute, all the reported results are from single runs.
We experiment with three different settings, summarized below:
\begin{enumerate}
    \item \textbf{Zero-shot setting.} Training is performed on English data only, and the model selection is done on the English development set. Results are reported in Table~\ref{table:parser-performance-zeroshot}.
    \item \textbf{Gold-data setting.} Training is performed on all the \massive{} data, that includes 51 languages. Model selection is performed averaging the accuracy on the multilingual development sets. Results are reported Table~\ref{table:parser-performance-gold}.
    \item \textbf{Synthetic data setting (\taf{}).} Training is performed on English and multilingual data that is synthetically generated via \taf{}. Results are reported in Table~\ref{table:parser-performance-synth}. Our entry based on this approach ranked 1st in the Zero-Shot Task of the MMNLU-22 Multilingual Semantic Parsing competition organized by Amazon and co-located with EMNLP 2022.\footnote{\url{https://mmnlu-22.github.io}}
\end{enumerate}

\begin{table}[ht]
%\small
\centering
\begin{tabular}{lS[table-format=2.2,detect-weight]S[table-format=2.2, detect-weight]H} % centered columns
\toprule
\textbf{Model} & \textbf{IA} & \textbf{EM} & \textbf{Steps} \\
\midrule
\byt5-small & 49.26 & 20.36 & 8k\\
\byt5-base  & 64.3 & 33.47 & 6k\\
\byt5-large & 66.53 & 28.43 & 10k\\
\byt5-xl    & 80.96 & 41.7 & 8k\\ % needs recomputation
\byt5-xxl   & 81.73 & 38.28 & 12k\\
\midrule
\mt5-small  & 51.75 & 17.59 & 18k\\
\mt5-base   & 55.91 & 17.73 & 20k\\
\mt5-large  & 67.23 & 25.14 & 18k\\
\mt5-xl     & 79.97 & 45.60 & 10k\\
\mt5-xxl    & {\B 82.44} & {\B 50.21} & 2k\\
% \midrule
% XLM-R-base  & 70.62 & 38.70 \\
\bottomrule
\end{tabular}
\caption{Zero-shot *T5 parsers performance when training on English only.}
\label{table:parser-performance-zeroshot}
\end{table}

\begin{table}[ht]
% \small
\centering
\begin{tabular}{lS[table-format=2.2, detect-weight]S[table-format=2.2, detect-weight]H} % centered columns
\toprule
\textbf{Model} & \textbf{IA} & \textbf{EM} & \textbf{Steps} \\
\midrule
\byt5-small & 85.59 & 66.60 & 30k\\
\byt5-base  & 85.93 & 67.54 & 30k\\
\byt5-large & 84.02 & 62.92 & 30k\\
\byt5-xl    & 87.01 & 68.29 & 20k\\
\byt5-xxl   & {\B 87.27} & {\B 68.66} & 30k\\
\midrule
\mt5-small  & 73.29 & 46.65 & 30k\\
\mt5-base   & 82.03 & 58.24 & 30k\\
\mt5-large  & 85.58 & 64.13 & 30k\\
\mt5-xl     & 87.24 & 68.47 & 10k\\
\mt5-xxl    & 86.79 & 63.33 & 5k\\
% \midrule
% XLM-R-base  & 85.10 & 63.69 \\
\bottomrule
\end{tabular}
\caption{*T5 parsers performance when training on all the available gold data.}
\label{table:parser-performance-gold}
\end{table}

%In the \textbf{zero-shot setting} (Table~\ref{table:parser-performance-zeroshot}), we train on English data only and do model selection on the English development data. In the \textbf{gold data setting} (Table~\ref{table:parser-performance-gold}), we train on all the available multilingual gold data and perform model selection averaging the accuracy on the multilingual development sets.
We can see a pattern that is common to all the experiments: at smaller sizes, \byt5{} has much better EM accuracy then the corresponding \mt5{} models. As stated in \citet{xu-etal-2020-end}, this may be explained by the fact that at these sizes less than 0.3\% of \byt5{} parameters are locked in embedding tables and a larger amount of dense parameters is updated during training. \mt5{} parameters are instead dominated by the embedding tables, which are updated less often than the dense layers. In addition to that, \byt5{}-large is worse than \byt5{}-base at span labeling, which is a word level task. Both our observations confirm the findings in~\citet{xu-etal-2020-end}.

\begin{table}[ht]
% \small
\centering
\begin{tabular}{lS[table-format=2.2, detect-weight]S[table-format=2.2, detect-weight]H} % centered columns
\toprule
\textbf{Model} & \textbf{IA} & \textbf{EM} & \textbf{Steps} \\
\midrule
\byt5-small & 83.32 & 59.32 & 25k\\
\byt5-base  & 84.59 & 61.24 & 25k\\
\byt5-large & 82.82 & 58.09 & 30k\\
\byt5-xl    & 85.90 & 62.98 & 15k\\
\byt5-xxl   & 86.48 & {\B 64.18} & 10k\\
\midrule
\mt5-small  & 73.64 & 43.19 & 30k\\
\mt5-base   & 80.79 & 51.76 & 30k\\
\mt5-large  & 83.99 & 57.43 & 30k\\
\mt5-xl     & 86.07 & 62.33 & 5k\\
\mt5-xxl    & {\B 86.69} & 62.49 & 20k\\
\bottomrule
\end{tabular}
\caption{*T5 parsers performance when training on English and synthetic \taf{} data.}
\label{table:parser-performance-synth}
\end{table}

In the \textbf{synthetic data setting} (Table~\ref{table:parser-performance-synth}),
IA almost matches the IA of models from the gold data setting. If we consider EM accuracy, we are only 5\% points behind the upper bound performance of the multilingually supervised -xxl models (see Table \ref{table:parser-performance-gold}). This indicates that synthetic data augmentation is a viable approach for the i18n of semantic parsers. Please refer to Table \ref{table:full-results} in the appendix for results on individual languages.

\section{Additional Experiments and Results}

\begin{figure*}[!ht]
  \centerline{
  \includegraphics[width=\textwidth]{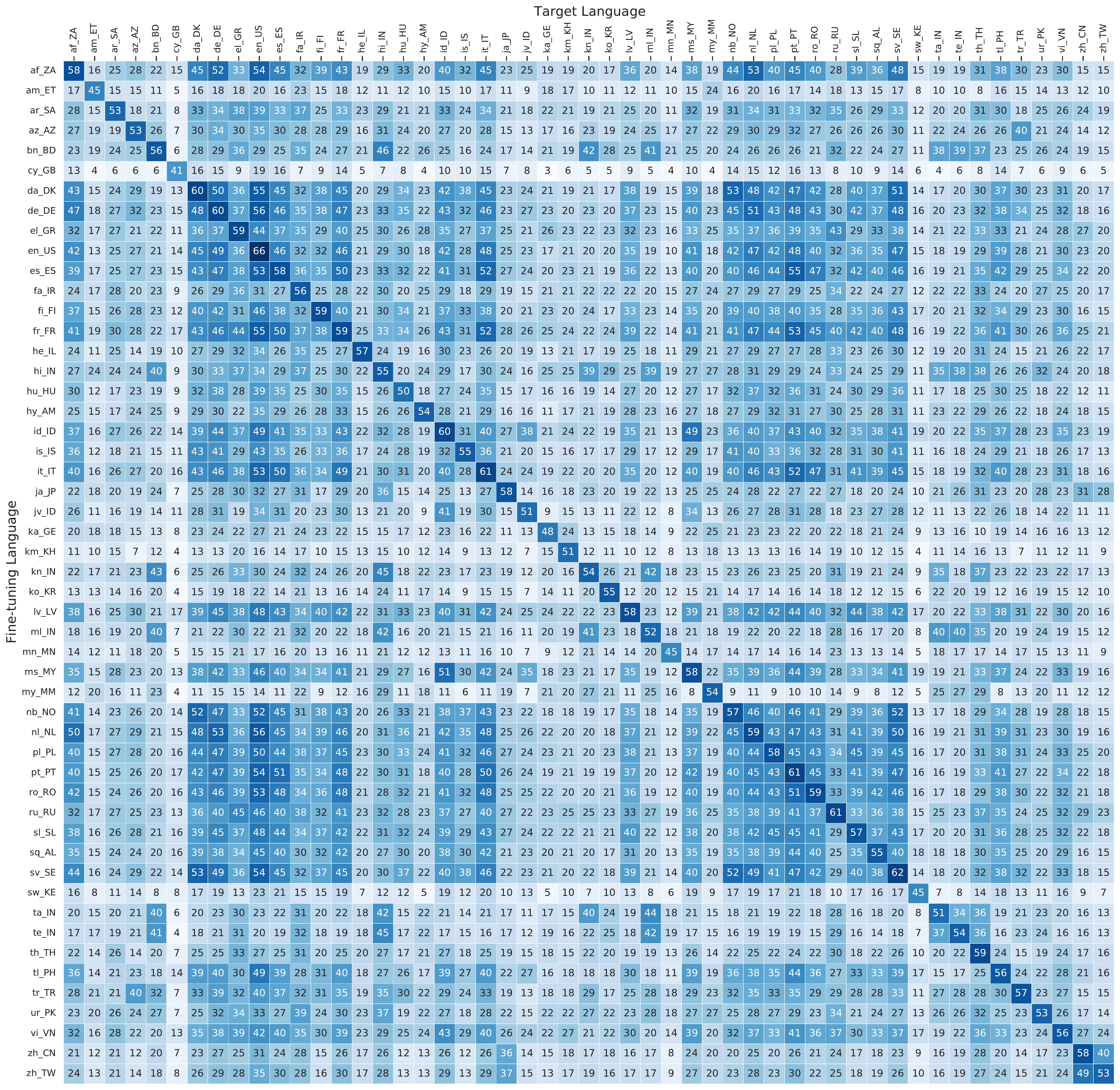}
  }
  \caption{Zero-shot EM accuracies of individual \byt5-base models fine-tuned on a single language (y-axis) and evaluated on dev sets from all languages (x-axis).}
\label{figure:cross-lingual-transfer}
\end{figure*}

In zero-shot evaluations, English is the most studied language given the availability of labeled data. Recent work has shown that this language may not be the best at cross-lingual transfer \cite{Turc2021RevisitingTP}. Since \massive{} provides training and test data for all its languages, we can evaluate the zero-shot performance of each language. We train $51$ \byt5-base model for a fixed number of steps (\si{1k} steps, $128$ batch size) and collect the results on the development sets in Figure~\ref{figure:cross-lingual-transfer}. By summing the EMs on rows we can understand how much a fine-tuning language (\textit{donor}) improves the others. If we sum over columns, we can see how much transfer a target language (\textit{receiver}) gets from the others. We report some statistics about best/worst donor/receiver languages in Table~\ref{table:best-worst-language}. Interestingly, English is not among the top donors, while it is the one that is being improved the most by other languages. We speculate that the better English LM representations may already have an intrinsic notion of semantic concepts that are then quickly individuated if supervision for such concepts is provided in other languages. From Figure~\ref{figure:cross-lingual-transfer}, we see that some languages (am, sw, km, cy) clearly need annotated data. We hope that this map could help prioritize data collection efforts.

\begin{figure}\centering
  \includegraphics[width=\columnwidth]{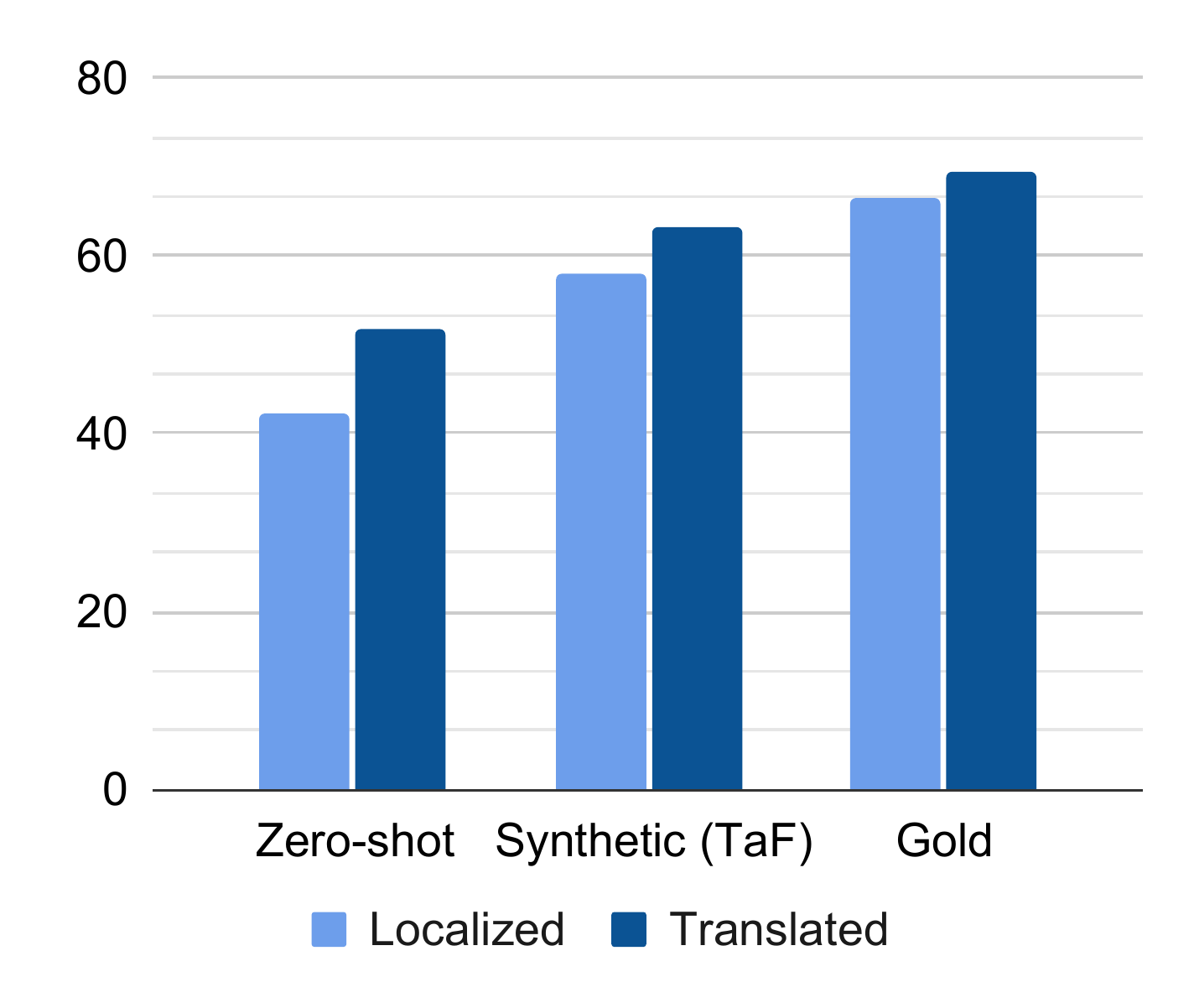}
  \caption{Differences in EM for an mT5-xxl model evaluated on queries of the test set that have been both translated and localized, vs only translated.}
  \label{figure:localized-vs-translated}
\end{figure}

\massive{} examples contain an interesting piece of metadata that indicates if an utterance has been translated and localized (i.e. original entities have been substituted with entities more culturally relevant for the target language), or translated only. We split the test sets in two parts according to this information and report in Figure~\ref{figure:localized-vs-translated} the EM accuracies of the same \mt5-xxl model. We examine the three data settings studied in this paper. Accuracies on \textit{localized} utterances are consistently lower. The performance difference in the synthetic data setting is relatively small but it still suggests that creating synthetic examples with entities that are \textit{local} to the target language may improve the robustness of the parser.

\begin{table}[ht]
\centering
\begin{tabular}{rc}
\toprule
                  & \textbf{Best to worst} \\
\midrule
\textbf{Donor}    & {\small fr, de, es, nl, pl, $\cdots$, mn, am, sw, km, cy} \\
\textbf{Receiver} & {\small en, de, pt, fr, sv, $\cdots$, zh, am, mn, sw, cy} \\
\bottomrule
\end{tabular}
\caption{Top-5 Best/worst donor/receiver.}
\label{table:best-worst-language}
\end{table}

In the appendix, we report the accuracy for each individual intent on the union of the test set examples from all languages (Table~\ref{table:intent-accuracies-full}). In Table~\ref{table:intent-accuracies-small}, we report the 6 intents with the lowest accuracy. Most examples belong to the \texttt{GENERAL\_QUIRKY} intent. The latter is likely a bucket intent covering all the utterances that are generic or out-of-domain (we could not find an exhaustive description of this intent in the SLURP dataset\cite{bastianelli-etal-2020-slurp}). The common parser mistake is to classify such queries as belonging to a more specific intent that can plausibly be associated with that query.

\begin{table}
\centering
\begin{tabular}{lS[table-format=2.1]S[table-format=4]}
\toprule
\textbf{Intent}             & \textbf{IA} & \textbf{Support} \\
\midrule
\texttt{GENERAL\_GREET}       & 19.6     & 51      \\
\texttt{MUSIC\_SETTINGS}      & 27.1     & 306     \\
\texttt{AUDIO\_VOLUME\_OTHER} & 54.9     & 306     \\
\texttt{GENERAL\_QUIRKY}      & 55.6     & 8619    \\
\texttt{IOT\_HUE\_LIGHTON}    & 61.4     & 153     \\
\texttt{MUSIC\_DISLIKENESS}   & 74.5     & 204     \\
\bottomrule
\end{tabular}
\caption{IA of the \byt5-xxl+\taf{} model for the lowest scoring intents (considering all languages).}
\label{table:intent-accuracies-small}
\end{table}

Finally, we compare our NMT translations of the training set with the corresponding gold translations produced by professional translators. We summarize the most interesting information in Table \ref{table:nmt-comparison-summary} (full comparison in Table \ref{table:verbatim-match-nmt-gold} included in the appendix). Indic languages (*\_IN and bn\_BD) have an higher average match than other languages. This may suggest that translations in these languages are more unambiguous or that translators may have relied on a MT during the translation task.

\begin{table}
\centering
\begin{tabular}{lS[table-format=2.1]}
\toprule
\textbf{Language sets} & \textbf{Avg Match} {(\%)}\\
\midrule
All languages  & 21.3     \\
All but Indic languages      & 17.3    \\
Indic languages & 50.8     \\
\bottomrule
\end{tabular}
\caption{Percentages of NMT translations matching human translations in \massive{} training set.}
\label{table:nmt-comparison-summary}
\end{table}

\section{Related Work}

\noindent\textbf{Multilingual models} are architecturally similar to monolingual transformer-based models but they are pretrained on multilingual corpora. These models include XLM \cite{LampleC2019}, XLM-R
\cite{Conneau20XLMR} and \mt5 \cite{xue2021mt5}, the multilingual version of T5~\cite{JMLR:v21:20-074}. They all use a subword vocabulary, a choice that may result in poor performance for languages with limited amount of data
\cite{wang-etal-2021-multi-view}.
Token-free models such as \byt5{}~\cite{xu-etal-2020-end}, Canine \cite{clark-etal-2022-canine} and Charformer~\cite{tay2022charformer} were designed to avoid this issue and have been applied to popular multilingual benchmarks obtaining state-of-the-art results.
In this work, we compare the multilinguality and the generative capabilities of \mt5{} and \byt5{} in a massively multilingual semantic parsing task.
\\\\
\noindent\textbf{Data augmentation} is the process of creating synthetic labeled data from available annotated examples. One approach in the multilinguality space is to translate annotated data in one language, e.g. English, to other languages. Neural machine translation is a strong baseline as it has been shown in recent cross-lingual evaluation benchmarks \cite{pmlr-v119-hu20b, ladhak-etal-2020-wikilingua}.
While translation works quite well for classification tasks where the label is at instance level, sequence tagging or parsing tasks require an annotation projection step because labels are at token level. Translate-and-align methods use bilingual word aligners, statistical \cite{brown-etal-1993-mathematics, vogel-etal-1996-hmm, och-ney-2000-improved, och03:asc}, and neural \cite{Schuster2019, chen-etal-2020-accurate, zenkel-etal-2020-end}.
More recent works removes this explicit alignment requirement~\cite{dong-lapata-2018-coarse, zhang-etal-2019-adansp, wiseman-etal-2018-learning}. In our work, we use a label projection method based on pretrained language models~\cite{nicosia-etal-2021-translate-fill} that reconstructs a full semantic parse from an utterance and a signature of the same parse.

\section{Conclusions}

In this paper, we evaluated \byt5{} and \mt5{}~\cite{xue2021mt5} models in a massively multilingual semantic parsing task, showing that \byt5 is particularly competitive at smaller sizes. We have provided a map of the cross-lingual transfer for all the languages in \massive{} and demonstrated that synthetic examples created with NMT are effective for building accurate semantic parsers.

\section*{Limitations}

This work uses seq2seq models as parsers. Different output formats can yield better or worse results as shown in \citet{paolini2021structured}. We do not focus on tweaking formats or on modeling improvements such as constrained decoding for a more faithful generation. We adopt a compact output representation that reduces the text the model has to generate (and hallucinations) and gives us competitive results. In the cross-lingual transfer experiments, we train each model for a small fixed number of steps. If we train for longer, the representations start to change significantly and cross-lingual performances vary quite unpredictably. We leave for the future an investigation of the learning dynamics in this setting and the design of possible remedies.

\bibliography{anthology,custom}

\begin{thebibliography}{34}
\expandafter\ifx\csname natexlab\endcsname\relax\def\natexlab#1{#1}\fi

\bibitem[{Artzi and Zettlemoyer(2011)}]{artzi2011bootstrapping}
Yoav Artzi and Luke Zettlemoyer. 2011.
\newblock Bootstrapping semantic parsers from conversations.
\newblock In \emph{Proceedings of the 2011 Conference on Empirical Methods in
  Natural Language Processing}, pages 421--432.

\bibitem[{Bastianelli et~al.(2020)Bastianelli, Vanzo, Swietojanski, and
  Rieser}]{bastianelli-etal-2020-slurp}
Emanuele Bastianelli, Andrea Vanzo, Pawel Swietojanski, and Verena Rieser.
  2020.
\newblock \href {https://doi.org/10.18653/v1/2020.emnlp-main.588} {{SLURP}: A
  spoken language understanding resource package}.
\newblock In \emph{Proceedings of the 2020 Conference on Empirical Methods in
  Natural Language Processing (EMNLP)}, pages 7252--7262, Online. Association
  for Computational Linguistics.

\bibitem[{Berant et~al.(2013)Berant, Chou, Frostig, and
  Liang}]{berant2013semantic}
Jonathan Berant, Andrew Chou, Roy Frostig, and Percy Liang. 2013.
\newblock Semantic parsing on freebase from question-answer pairs.
\newblock In \emph{Proceedings of the 2013 conference on empirical methods in
  natural language processing}, pages 1533--1544.

\bibitem[{Brown et~al.(1993)Brown, Della~Pietra, Della~Pietra, and
  Mercer}]{brown-etal-1993-mathematics}
Peter~F. Brown, Stephen~A. Della~Pietra, Vincent~J. Della~Pietra, and Robert~L.
  Mercer. 1993.
\newblock \href {https://www.aclweb.org/anthology/J93-2003} {The mathematics of
  statistical machine translation: Parameter estimation}.
\newblock \emph{Computational Linguistics}, 19(2):263--311.

\bibitem[{Chen et~al.(2020)Chen, Liu, Chen, Jiang, and
  Liu}]{chen-etal-2020-accurate}
Yun Chen, Yang Liu, Guanhua Chen, Xin Jiang, and Qun Liu. 2020.
\newblock \href {https://doi.org/10.18653/v1/2020.emnlp-main.42} {Accurate word
  alignment induction from neural machine translation}.
\newblock In \emph{Proceedings of the 2020 Conference on Empirical Methods in
  Natural Language Processing (EMNLP)}, pages 566--576, Online. Association for
  Computational Linguistics.

\bibitem[{Clark et~al.(2022)Clark, Garrette, Turc, and
  Wieting}]{clark-etal-2022-canine}
Jonathan~H. Clark, Dan Garrette, Iulia Turc, and John Wieting. 2022.
\newblock \href {https://doi.org/10.1162/tacl_a_00448} {Canine: Pre-training an
  efficient tokenization-free encoder for language representation}.
\newblock \emph{Transactions of the Association for Computational Linguistics},
  10:73--91.

\bibitem[{Conneau et~al.(2020)Conneau, Khandelwal, Goyal, Chaudhary, Wenzek,
  Guzm{\'a}n, Grave, Ott, Zettlemoyer, and Stoyanov}]{Conneau20XLMR}
Alexis Conneau, Kartikay Khandelwal, Naman Goyal, Vishrav Chaudhary, Guillaume
  Wenzek, Francisco Guzm{\'a}n, Edouard Grave, Myle Ott, Luke Zettlemoyer, and
  Veselin Stoyanov. 2020.
\newblock \href {https://doi.org/10.18653/v1/2020.acl-main.747} {Unsupervised
  cross-lingual representation learning at scale}.
\newblock In \emph{Proceedings of the 58th Annual Meeting of the Association
  for Computational Linguistics}, pages 8440--8451, Online. Association for
  Computational Linguistics.

\bibitem[{Ding et~al.(2022)Ding, Hu, Bing, Aljunied, Joty, Si, and
  Miao}]{ding-etal-2022-globalwoz}
Bosheng Ding, Junjie Hu, Lidong Bing, Mahani Aljunied, Shafiq Joty, Luo Si, and
  Chunyan Miao. 2022.
\newblock \href {https://doi.org/10.18653/v1/2022.acl-long.115}
  {{G}lobal{W}o{Z}: Globalizing {M}ulti{W}o{Z} to develop multilingual
  task-oriented dialogue systems}.
\newblock In \emph{Proceedings of the 60th Annual Meeting of the Association
  for Computational Linguistics (Volume 1: Long Papers)}, pages 1639--1657,
  Dublin, Ireland. Association for Computational Linguistics.

\bibitem[{Dong and Lapata(2018)}]{dong-lapata-2018-coarse}
Li~Dong and Mirella Lapata. 2018.
\newblock \href {https://doi.org/10.18653/v1/P18-1068} {Coarse-to-fine decoding
  for neural semantic parsing}.
\newblock In \emph{Proceedings of the 56th Annual Meeting of the Association
  for Computational Linguistics (Volume 1: Long Papers)}, pages 731--742,
  Melbourne, Australia. Association for Computational Linguistics.

\bibitem[{Dukes(2014)}]{dukes2014semeval}
Kais Dukes. 2014.
\newblock Semeval-2014 task 6: Supervised semantic parsing of robotic spatial
  commands.
\newblock In \emph{SemEval@ COLING}, pages 45--53.

\bibitem[{FitzGerald et~al.(2022)FitzGerald, Hench, Peris, Mackie, Rottmann,
  Sanchez, Nash, Urbach, Kakarala, Singh, Ranganath, Crist, Britan, Leeuwis,
  Tur, and Natarajan}]{massive}
Jack FitzGerald, Christopher Hench, Charith Peris, Scott Mackie, Kay Rottmann,
  Ana Sanchez, Aaron Nash, Liam Urbach, Vishesh Kakarala, Richa Singh, Swetha
  Ranganath, Laurie Crist, Misha Britan, Wouter Leeuwis, Gokhan Tur, and Prem
  Natarajan. 2022.
\newblock \href {https://doi.org/10.48550/ARXIV.2204.08582} {Massive: A
  1m-example multilingual natural language understanding dataset with 51
  typologically-diverse languages}.

\bibitem[{Hu et~al.(2020)Hu, Ruder, Siddhant, Neubig, Firat, and
  Johnson}]{pmlr-v119-hu20b}
Junjie Hu, Sebastian Ruder, Aditya Siddhant, Graham Neubig, Orhan Firat, and
  Melvin Johnson. 2020.
\newblock \href {http://proceedings.mlr.press/v119/hu20b.html} {{XTREME}: A
  massively multilingual multi-task benchmark for evaluating cross-lingual
  generalisation}.
\newblock In \emph{Proceedings of the 37th International Conference on Machine
  Learning}, volume 119 of \emph{Proceedings of Machine Learning Research},
  pages 4411--4421. PMLR.

\bibitem[{Ladhak et~al.(2020)Ladhak, Durmus, Cardie, and
  McKeown}]{ladhak-etal-2020-wikilingua}
Faisal Ladhak, Esin Durmus, Claire Cardie, and Kathleen McKeown. 2020.
\newblock \href {https://doi.org/10.18653/v1/2020.findings-emnlp.360}
  {{W}iki{L}ingua: A new benchmark dataset for cross-lingual abstractive
  summarization}.
\newblock In \emph{Findings of the Association for Computational Linguistics:
  EMNLP 2020}, pages 4034--4048, Online. Association for Computational
  Linguistics.

\bibitem[{Lample and Conneau(2019)}]{LampleC2019}
Guillaume Lample and Alexis Conneau. 2019.
\newblock \href {http://arxiv.org/abs/1901.07291} {Cross-lingual language model
  pretraining}.
\newblock \emph{CoRR}, abs/1901.07291.

\bibitem[{Li et~al.(2021)Li, Arora, Chen, Gupta, Gupta, and
  Mehdad}]{li-etal-2021-mtop}
Haoran Li, Abhinav Arora, Shuohui Chen, Anchit Gupta, Sonal Gupta, and Yashar
  Mehdad. 2021.
\newblock \href {https://www.aclweb.org/anthology/2021.eacl-main.257} {{MTOP}:
  A comprehensive multilingual task-oriented semantic parsing benchmark}.
\newblock In \emph{Proceedings of the 16th Conference of the European Chapter
  of the Association for Computational Linguistics: Main Volume}, pages
  2950--2962, Online. Association for Computational Linguistics.

\bibitem[{Lin et~al.(2021)Lin, Madotto, Winata, Xu, Jiang, Hu, Shi, and
  Fung}]{lin2021bitod}
Zhaojiang Lin, Andrea Madotto, Genta~Indra Winata, Peng Xu, Feijun Jiang,
  Yuxiang Hu, Chen Shi, and Pascale Fung. 2021.
\newblock \href {https://openreview.net/forum?id=dA2Q8CfmGpp} {Bitod: A
  bilingual multi-domain dataset for task-oriented dialogue modeling}.
\newblock In \emph{Thirty-fifth Conference on Neural Information Processing
  Systems Datasets and Benchmarks Track (Round 1)}.

\bibitem[{Majewska et~al.(2022)Majewska, Razumovskaia, Ponti, Vulić, and
  Korhonen}]{https://doi.org/10.48550/arxiv.2201.13405}
Olga Majewska, Evgeniia Razumovskaia, Edoardo~Maria Ponti, Ivan Vulić, and
  Anna Korhonen. 2022.
\newblock \href {https://doi.org/10.48550/ARXIV.2201.13405} {Cross-lingual
  dialogue dataset creation via outline-based generation}.

\bibitem[{Moradshahi et~al.(2020)Moradshahi, Campagna, Semnani, Xu, and
  Lam}]{moradshahi-etal-2020-localizing}
Mehrad Moradshahi, Giovanni Campagna, Sina Semnani, Silei Xu, and Monica Lam.
  2020.
\newblock \href {https://doi.org/10.18653/v1/2020.emnlp-main.481} {Localizing
  open-ontology {QA} semantic parsers in a day using machine translation}.
\newblock In \emph{Proceedings of the 2020 Conference on Empirical Methods in
  Natural Language Processing (EMNLP)}, pages 5970--5983, Online. Association
  for Computational Linguistics.

\bibitem[{Nicosia et~al.(2021)Nicosia, Qu, and
  Altun}]{nicosia-etal-2021-translate-fill}
Massimo Nicosia, Zhongdi Qu, and Yasemin Altun. 2021.
\newblock \href {https://doi.org/10.18653/v1/2021.findings-emnlp.279}
  {{T}ranslate {\&} {F}ill: {I}mproving zero-shot multilingual semantic parsing
  with synthetic data}.
\newblock In \emph{Findings of the Association for Computational Linguistics:
  EMNLP 2021}, pages 3272--3284, Punta Cana, Dominican Republic. Association
  for Computational Linguistics.

\bibitem[{Och and Ney(2000)}]{och-ney-2000-improved}
Franz~Josef Och and Hermann Ney. 2000.
\newblock \href {https://doi.org/10.3115/1075218.1075274} {Improved statistical
  alignment models}.
\newblock In \emph{Proceedings of the 38th Annual Meeting of the Association
  for Computational Linguistics}, pages 440--447, Hong Kong. Association for
  Computational Linguistics.

\bibitem[{Och and Ney(2003)}]{och03:asc}
Franz~Josef Och and Hermann Ney. 2003.
\newblock A systematic comparison of various statistical alignment models.
\newblock \emph{Computational Linguistics}, 29(1):19--51.

\bibitem[{Paolini et~al.(2021)Paolini, Athiwaratkun, Krone, Ma, Achille,
  ANUBHAI, dos Santos, Xiang, and Soatto}]{paolini2021structured}
Giovanni Paolini, Ben Athiwaratkun, Jason Krone, Jie Ma, Alessandro Achille,
  RISHITA ANUBHAI, Cicero~Nogueira dos Santos, Bing Xiang, and Stefano Soatto.
  2021.
\newblock \href {https://openreview.net/forum?id=US-TP-xnXI} {Structured
  prediction as translation between augmented natural languages}.
\newblock In \emph{International Conference on Learning Representations}.

\bibitem[{Raffel et~al.(2020)Raffel, Shazeer, Roberts, Lee, Narang, Matena,
  Zhou, Li, and Liu}]{JMLR:v21:20-074}
Colin Raffel, Noam Shazeer, Adam Roberts, Katherine Lee, Sharan Narang, Michael
  Matena, Yanqi Zhou, Wei Li, and Peter~J. Liu. 2020.
\newblock \href {http://jmlr.org/papers/v21/20-074.html} {Exploring the limits
  of transfer learning with a unified text-to-text transformer}.
\newblock \emph{Journal of Machine Learning Research}, 21(140):1--67.

\bibitem[{Schuster et~al.(2019)Schuster, Gupta, Shah, and Lewis}]{Schuster2019}
Sebastian Schuster, Sonal Gupta, Rushin Shah, and Mike Lewis. 2019.
\newblock \href {https://doi.org/10.18653/v1/N19-1380} {Cross-lingual transfer
  learning for multilingual task oriented dialog}.
\newblock In \emph{Proceedings of the 2019 Conference of the North {A}merican
  Chapter of the Association for Computational Linguistics: Human Language
  Technologies, Volume 1 (Long and Short Papers)}, pages 3795--3805.
  Association for Computational Linguistics.

\bibitem[{Sherborne et~al.(2020)Sherborne, Xu, and
  Lapata}]{sherborne-etal-2020-bootstrapping}
Tom Sherborne, Yumo Xu, and Mirella Lapata. 2020.
\newblock \href {https://doi.org/10.18653/v1/2020.findings-emnlp.45}
  {Bootstrapping a crosslingual semantic parser}.
\newblock In \emph{Findings of the Association for Computational Linguistics:
  EMNLP 2020}, pages 499--517, Online. Association for Computational
  Linguistics.

\bibitem[{Tay et~al.(2022)Tay, Tran, Ruder, Gupta, Chung, Bahri, Qin,
  Baumgartner, Yu, and Metzler}]{tay2022charformer}
Yi~Tay, Vinh~Q. Tran, Sebastian Ruder, Jai Gupta, Hyung~Won Chung, Dara Bahri,
  Zhen Qin, Simon Baumgartner, Cong Yu, and Donald Metzler. 2022.
\newblock \href {https://openreview.net/forum?id=JtBRnrlOEFN} {Charformer: Fast
  character transformers via gradient-based subword tokenization}.
\newblock In \emph{International Conference on Learning Representations}.

\bibitem[{Turc et~al.(2021)Turc, Lee, Eisenstein, Chang, and
  Toutanova}]{Turc2021RevisitingTP}
Iulia Turc, Kenton Lee, Jacob Eisenstein, Ming-Wei Chang, and Kristina
  Toutanova. 2021.
\newblock Revisiting the primacy of english in zero-shot cross-lingual
  transfer.
\newblock \emph{ArXiv}, abs/2106.16171.

\bibitem[{Vogel et~al.(1996)Vogel, Ney, and Tillmann}]{vogel-etal-1996-hmm}
Stephan Vogel, Hermann Ney, and Christoph Tillmann. 1996.
\newblock \href {https://www.aclweb.org/anthology/C96-2141} {{HMM}-based word
  alignment in statistical translation}.
\newblock In \emph{{COLING} 1996 Volume 2: The 16th International Conference on
  Computational Linguistics}.

\bibitem[{Wang et~al.(2021)Wang, Ruder, and Neubig}]{wang-etal-2021-multi-view}
Xinyi Wang, Sebastian Ruder, and Graham Neubig. 2021.
\newblock \href {https://doi.org/10.18653/v1/2021.naacl-main.40} {Multi-view
  subword regularization}.
\newblock In \emph{Proceedings of the 2021 Conference of the North American
  Chapter of the Association for Computational Linguistics: Human Language
  Technologies}, pages 473--482, Online. Association for Computational
  Linguistics.

\bibitem[{Wiseman et~al.(2018)Wiseman, Shieber, and
  Rush}]{wiseman-etal-2018-learning}
Sam Wiseman, Stuart Shieber, and Alexander Rush. 2018.
\newblock \href {https://doi.org/10.18653/v1/D18-1356} {Learning neural
  templates for text generation}.
\newblock In \emph{Proceedings of the 2018 Conference on Empirical Methods in
  Natural Language Processing}, pages 3174--3187, Brussels, Belgium.
  Association for Computational Linguistics.

\bibitem[{Xu et~al.(2020)Xu, Haider, and Mansour}]{xu-etal-2020-end}
Weijia Xu, Batool Haider, and Saab Mansour. 2020.
\newblock \href {https://doi.org/10.18653/v1/2020.emnlp-main.410} {End-to-end
  slot alignment and recognition for cross-lingual {NLU}}.
\newblock In \emph{Proceedings of the 2020 Conference on Empirical Methods in
  Natural Language Processing (EMNLP)}, pages 5052--5063, Online. Association
  for Computational Linguistics.

\bibitem[{Xue et~al.(2021)Xue, Constant, Roberts, Kale, Al-Rfou, Siddhant,
  Barua, and Raffel}]{xue2021mt5}
Linting Xue, Noah Constant, Adam Roberts, Mihir Kale, Rami Al-Rfou, Aditya
  Siddhant, Aditya Barua, and Colin Raffel. 2021.
\newblock \href {http://arxiv.org/abs/2010.11934} {mt5: A massively
  multilingual pre-trained text-to-text transformer}.

\bibitem[{Zenkel et~al.(2020)Zenkel, Wuebker, and
  DeNero}]{zenkel-etal-2020-end}
Thomas Zenkel, Joern Wuebker, and John DeNero. 2020.
\newblock \href {https://doi.org/10.18653/v1/2020.acl-main.146} {End-to-end
  neural word alignment outperforms {GIZA}++}.
\newblock In \emph{Proceedings of the 58th Annual Meeting of the Association
  for Computational Linguistics}, pages 1605--1617, Online. Association for
  Computational Linguistics.

\bibitem[{Zhang et~al.(2019)Zhang, He, Liu, and Zhao}]{zhang-etal-2019-adansp}
Xiang Zhang, Shizhu He, Kang Liu, and Jun Zhao. 2019.
\newblock \href {https://doi.org/10.18653/v1/P19-1418} {{A}da{NSP}:
  Uncertainty-driven adaptive decoding in neural semantic parsing}.
\newblock In \emph{Proceedings of the 57th Annual Meeting of the Association
  for Computational Linguistics}, pages 4265--4270, Florence, Italy.
  Association for Computational Linguistics.

\end{thebibliography}
\bibliographystyle{acl_natbib}

\clearpage
\newpage
\appendix

\section{Comparing NMT with Gold Translations}

In Table~\ref{table:verbatim-match-nmt-gold}, we compare how many times the NMT translated utterances match the gold translations produced by professional translators. We restrict the match to utterances that have been translated and not localized in the target language, since NMT cannot perform the localization step. In addition, we preprocess all compared utterances with unicode normalization, we strip whitespaces and punctuation. In general, indic locales have higher match rates compared to other locales. Please also note that we translate English to pt\_BR (Brazilian Portuguese) and this explains the low match for pt\_PT.

\section{Intent Accuracy Performance}

In Table~\ref{table:intent-accuracies-full}, we report the accuracy for each individual intent on the union of the test set examples from all languages using \byt5{}-xxl + \taf{}.

\begin{table*}
\begin{minipage}[t]{\columnwidth}
\small
\centering
\begin{tabular}{>{\ttfamily}cS[table-format=2.1,table-text-alignment=right,table-number-alignment=right]S[table-format=4,table-text-alignment=right,table-number-alignment=right]S[table-format=5, table-text-alignment=center]}
\toprule
 & \multicolumn{2}{c}{\textbf{NMT vs Gold}} & {\textbf{Non-localized}} \\
 & \multicolumn{2}{c}{\textbf{Translations Matches}} & {\textbf{sentences}} \\
\textbf{Language} & {(\%)} & {(\#)} & {(\#)} \\
\midrule
kn\_IN & 68.7 & 6524 & 9497  \\
te\_IN & 54.1 & 4841 & 8941  \\
bn\_BD & 52.6 & 4458 & 8471  \\
ta\_IN & 48.3 & 4301 & 8898  \\
hi\_IN & 46.5 & 4101 & 8827  \\
nl\_NL & 38.5 & 3878 & 10070 \\
fr\_FR & 36.0 & 3736 & 10385 \\
ml\_IN & 34.7 & 2985 & 8607  \\
tl\_PH & 34.0 & 3397 & 10000 \\
af\_ZA & 32.8 & 3160 & 9640  \\
tr\_TR & 32.1 & 2998 & 9330  \\
sw\_KE & 26.1 & 2336 & 8965  \\
sv\_SE & 25.9 & 2465 & 9504  \\
nb\_NO & 23.8 & 2402 & 10083 \\
vi\_VN & 21.6 & 2000 & 9255  \\
ms\_MY & 21.6 & 1880 & 8702  \\
jv\_ID & 21.1 & 1947 & 9208  \\
pl\_PL & 21.0 & 2017 & 9618  \\
da\_DK & 20.4 & 1933 & 9470  \\
id\_ID & 20.4 & 1882 & 9227  \\
es\_ES & 19.5 & 1876 & 9596  \\
zh\_CN & 19.0 & 1661 & 8727  \\
zh\_TW & 18.2 & 1638 & 8976  \\
it\_IT & 17.9 & 1596 & 8916  \\
fi\_FI & 17.5 & 1669 & 9558  \\
ru\_RU & 17.4 & 1550 & 8912  \\
hy\_AM & 16.9 & 1809 & 10707 \\
is\_IS & 16.1 & 1491 & 9270  \\
km\_KH & 16.1 & 1491 & 9276  \\
cy\_GB & 15.9 & 1578 & 9936  \\
sl\_SL & 14.7 & 1313 & 8913  \\
am\_ET & 14.6 & 1267 & 8658  \\
hu\_HU & 14.5 & 1331 & 9198  \\
ur\_PK & 14.4 & 1260 & 8761  \\
de\_DE & 14.2 & 1422 & 9992  \\
lv\_LV & 12.4 & 1071 & 8650  \\
he\_IL & 12.3 & 1123 & 9159  \\
sq\_AL & 12.2 & 1035 & 8460  \\
az\_AZ & 12.1 & 1102 & 9081  \\
th\_TH & 11.7 & 1041 & 8894  \\
ro\_RO & 10.9 & 1001 & 9197  \\
el\_GR & 10.5 & 934  & 8879  \\
pt\_PT & 9.9  & 934  & 9392  \\
ar\_SA & 9.9  & 871  & 8814  \\
mn\_MN & 8.9  & 785  & 8826  \\
fa\_IR & 8.3  & 718  & 8686  \\
ja\_JP & 7.4  & 704  & 9487  \\
ka\_GE & 7.4  & 701  & 9528  \\
ko\_KR & 3.9  & 341  & 8804  \\
my\_MM & 2.0  & 171  & 8765  \\
\bottomrule
\end{tabular}
\caption{Number of verbatim matches between Gold translation and NMT translations.}
\label{table:verbatim-match-nmt-gold}

\end{minipage}\hfill % maximize the horizontal separation
\begin{minipage}[t]{\columnwidth}
\small
\centering
\begin{tabular}{lS[table-format=2.1]S[table-format=4]}
\toprule
\textbf{Intent}             & \textbf{IA} & \textbf{Support} \\
\midrule
\texttt{GENERAL\_GREET}            & 19.6     & 51      \\
\texttt{MUSIC\_SETTINGS}           & 27.1     & 306     \\
\texttt{AUDIO\_VOLUME\_OTHER}      & 54.9     & 306     \\
\texttt{GENERAL\_QUIRKY}           & 55.6     & 8619    \\
\texttt{IOT\_HUE\_LIGHTON}         & 61.4     & 153     \\
\texttt{MUSIC\_DISLIKENESS}        & 74.5     & 204     \\
\texttt{DATETIME\_CONVERT}         & 75.6     & 765     \\
\texttt{IOT\_WEMO\_ON}             & 76.3     & 510     \\
\texttt{PLAY\_AUDIOBOOK}           & 78.0     & 2091    \\
\texttt{TRANSPORT\_QUERY}          & 78.1     & 2601    \\
\texttt{RECOMMENDATION\_EVENTS}    & 78.3     & 2193    \\
\texttt{RECOMMENDATION\_MOVIES}    & 79.2     & 1020    \\
\texttt{CALENDAR\_QUERY}           & 80.6     & 6426    \\
\texttt{QA\_FACTOID}               & 82.4     & 7191    \\
\texttt{IOT\_HUE\_LIGHTUP}         & 82.5     & 1377    \\
\texttt{LISTS\_QUERY}              & 82.6     & 2601    \\
\texttt{AUDIO\_VOLUME\_UP}         & 83.0     & 663     \\
\texttt{SOCIAL\_QUERY}             & 83.9     & 1275    \\
\texttt{MUSIC\_QUERY}              & 84.0     & 1785    \\
\texttt{EMAIL\_ADDCONTACT}         & 84.5     & 612     \\
\texttt{MUSIC\_LIKENESS}           & 84.7     & 1836    \\
\texttt{EMAIL\_QUERYCONTACT}       & 84.8     & 1326    \\
\texttt{TAKEAWAY\_QUERY}           & 85.0     & 1785    \\
\texttt{LISTS\_CREATEORADD}        & 85.6     & 1989    \\
\texttt{QA\_DEFINITION}            & 86.3     & 2907    \\
\texttt{LISTS\_REMOVE}             & 86.3     & 2652    \\
\texttt{COOKING\_RECIPE}           & 86.6     & 3672    \\
\texttt{NEWS\_QUERY}               & 86.9     & 6324    \\
\texttt{PLAY\_MUSIC}               & 87.1     & 8976    \\
\texttt{TAKEAWAY\_ORDER}           & 87.3     & 1122    \\
\texttt{IOT\_HUE\_LIGHTDIM}        & 87.4     & 1071    \\
\texttt{PLAY\_PODCASTS}            & 87.6     & 3213    \\
\texttt{PLAY\_GAME}                & 87.7     & 1785    \\
\texttt{ALARM\_SET}                & 89.5     & 2091    \\
\texttt{PLAY\_RADIO}               & 90.0     & 3672    \\
\texttt{CALENDAR\_SET}             & 90.2     & 10659   \\
\texttt{RECOMMENDATION\_LOCATIONS} & 90.4     & 1581    \\
\texttt{QA\_MATHS}                 & 90.7     & 1275    \\
\texttt{AUDIO\_VOLUME\_DOWN}       & 90.7     & 561     \\
\texttt{SOCIAL\_POST}              & 91.1     & 4131    \\
\texttt{IOT\_WEMO\_OFF}            & 91.3     & 918     \\
\texttt{AUDIO\_VOLUME\_MUTE}       & 91.7     & 1632    \\
\texttt{ALARM\_QUERY}              & 91.8     & 1734    \\
\texttt{GENERAL\_JOKE}             & 92.0     & 969     \\
\texttt{EMAIL\_QUERY}              & 93.0     & 6069    \\
\texttt{TRANSPORT\_TICKET}         & 93.1     & 1785    \\
\texttt{CALENDAR\_REMOVE}          & 93.4     & 3417    \\
\texttt{EMAIL\_SENDEMAIL}          & 94.0     & 5814    \\
\texttt{IOT\_CLEANING}             & 94.2     & 1326    \\
\texttt{WEATHER\_QUERY}            & 94.6     & 7956    \\
\texttt{IOT\_HUE\_LIGHTOFF}        & 94.8     & 2193    \\
\texttt{TRANSPORT\_TAXI}           & 95.3     & 1173    \\
\texttt{IOT\_HUE\_LIGHTCHANGE}     & 95.4     & 1836    \\
\texttt{ALARM\_REMOVE}             & 95.5     & 1071    \\
\texttt{QA\_STOCK}                 & 95.6     & 1326    \\
\texttt{DATETIME\_QUERY}           & 95.8     & 4488    \\
\texttt{TRANSPORT\_TRAFFIC}        & 96.3     & 765     \\
\texttt{QA\_CURRENCY}              & 96.6     & 1989    \\
\texttt{IOT\_COFFEE}               & 97.9     & 1836    \\
\bottomrule
\end{tabular}
\caption{IA of the \byt5-xxl+\taf{} model for all intents (all languages).}
\label{table:intent-accuracies-full}

  \end{minipage}
\end{table*}

\section{Performance on all Languages}

In Table~\ref{table:full-results}, we report Exact Match on all the 51 languages, for the three different experimental setups described in Section~\ref{sec:experiments}, across two models (\mt5 and \byt5) and two model sizes (base and xxl).

\begin{table*}[tb!]
\small
\centering
\begin{tabular}{cS[table-format=2.1, detect-weight]S[table-format=2.1, detect-weight]S[table-format=2.1, detect-weight]S[table-format=2.1, detect-weight]|S[table-format=2.1, detect-weight]S[table-format=2.1, detect-weight]S[table-format=2.1, detect-weight]S[table-format=2.1, detect-weight]|S[table-format=2.1, detect-weight]S[table-format=2.1, detect-weight]S[table-format=2.1, detect-weight]S[table-format=2.1, detect-weight]}
\toprule
         & \multicolumn{4}{c}{\textbf{Zero Shot}} & \multicolumn{4}{c}{\textbf{Synthetic (\taf{})}}       & \multicolumn{4}{c}{\textbf{Gold}}      \\
 & \multicolumn{2}{c}{\textbf{base}} & \multicolumn{2}{c}{\textbf{xxl}} & \multicolumn{2}{c}{\textbf{base}} & \multicolumn{2}{c}{\textbf{xxl}} & \multicolumn{2}{c}{\textbf{base}} & \multicolumn{2}{c}{\textbf{xxl}} \\
\textbf{Language} & \textbf{\mt5{}}   & \textbf{\byt5{}}  & \textbf{\mt5{}}   & \textbf{\byt5{}}  & \textbf{\mt5{}}   & \textbf{\byt5{}}  & \textbf{\mt5{}}   & \textbf{\byt5{}}  & \textbf{\mt5{}}   & \textbf{\byt5{}}  & \textbf{\mt5{}}   & \textbf{\byt5{}}  \\
\midrule
\texttt{af\_ZA}   & 21.6  & {\B 51.1}  & 58.0  & {\B 59.7}  & 53.7 & {\B 64.7} & 65.6 & {\B 66.8} & 59.4 & {\B 68.5} & 65.9 & {\B 69.3} \\
\texttt{am\_ET}   & 4.7   & {\B 15.9}  & {\B 40.7}  & 22.0  & 40.8 & {\B 54.4} & {\B 61.2} & 61.0 & 48.7 & {\B 61.3} & 62.0 & {\B 65.8} \\
\texttt{ar\_SA}   & 14.6  & {\B 27.8}  & {\B 43.6}  & 23.3  & 45.9 & {\B 56.1} & 60.1 & {\B 60.5} & 52.3 & {\B 64.7} & 61.1 & {\B 66.0} \\
\texttt{az\_AZ}   & 8.9   & {\B 31.2}  & {\B 41.8}  & 34.0  & 46.4 & {\B 61.6} & 61.9 & {\B 63.6} & 57.0 & {\B 69.0} & 62.6 & {\B 69.6} \\
\texttt{bn\_BD}   & 10.8  & {\B 19.5}  & {\B 45.9}  & 25.3  & 51.0 & {\B 62.1} & 64.3 & {\B 65.6} & 57.6 & {\B 67.6} & 64.6 & {\B 69.5} \\
\texttt{cy\_GB}   & 5.9   & {\B 16.4}  & {\B 42.8}  & 40.2  & 35.7 & {\B 56.1} & 61.5 & {\B 64.2} & 42.1 & {\B 65.3} & 61.4 & {\B 69.2} \\
\texttt{da\_DK}   & 30.2  & {\B 53.1}  & {\B 60.9}  & 54.2  & 57.8 & {\B 67.5} & 67.3 & {\B 68.7} & 64.4 & {\B 71.7} & 67.9 & {\B 71.3} \\
\texttt{de\_DE}   & 28.3  & {\B 55.3}  & {\B 59.8}  & 59.5  & 60.2 & {\B 67.8} & 67.5 & {\B 68.8} & 64.1 & {\B 70.4} & 68.0 & {\B 70.2} \\
\texttt{el\_GR}   & 17.4  & {\B 31.5}  & {\B 57.2}  & 27.9  & 55.5 & {\B 64.2} & 65.5 & {\B 66.6} & 62.0 & {\B 68.3} & 66.6 & {\B 68.7} \\
\texttt{en\_US}   & 65.5  & {\B 72.2}  & {\B 74.0}  & 73.3  & 68.5 & {\B 72.6} & {\B 73.7} & 73.0 & 68.9 & {\B 72.7} & {\B 73.3} & 72.6 \\
\texttt{es\_ES}   & 26.1  & {\B 50.8}  & {\B 55.6}  & 52.2  & 58.7 & {\B 65.1} & 65.0 & {\B 65.9} & 61.1 & {\B 67.2} & 65.9 & {\B 66.2} \\
\texttt{fa\_IR}   & 17.6  & {\B 32.8}  & {\B 54.4}  & 24.0  & 54.9 & {\B 62.2} & 63.2 & {\B 64.4} & 59.9 & {\B 69.1} & 63.4 & {\B 69.7} \\
\texttt{fi\_FI}   & 16.3  & {\B 36.9}  & {\B 52.5}  & 47.4  & 51.2 & {\B 65.9} & 65.6 & {\B 68.2} & 59.4 & {\B 71.1} & 66.8 & {\B 71.5} \\
\texttt{fr\_FR}   & 29.9  & {\B 53.5}  & {\B 58.5}  & 54.3  & 59.3 & {\B 64.4} & 65.1 & {\B 65.6} & 62.3 & {\B 66.5} & 65.8 & {\B 67.2} \\
\texttt{he\_IL}   & 9.7   & {\B 21.0}  & {\B 40.4}  & 24.0  & 50.1 & {\B 59.4} & 61.0 & {\B 63.2} & 57.5 & {\B 67.3} & 62.3 & {\B 68.4} \\
\texttt{hi\_IN}   & 14.1  & {\B 26.3}  & {\B 52.9}  & 26.2  & 54.4 & {\B 62.6} & 64.2 & {\B 64.4} & 59.3 & {\B 66.5} & 64.5 & {\B 67.2} \\
\texttt{hu\_HU}   & 17.5  & {\B 33.5}  & {\B 45.3}  & 32.9  & 51.8 & {\B 62.2} & 64.2 & 64.2      & 58.2 & {\B 68.5} & 65.2 & {\B 69.5} \\
\texttt{hy\_AM}   & 11.7  & {\B 20.5}  & {\B 44.6}  & 24.7  & 49.8 & {\B 58.4} & 60.3 & {\B 62.2} & 57.8 & {\B 67.7} & 61.7 & {\B 68.9} \\
\texttt{id\_ID}   & 24.1  & {\B 48.3}  & 58.6  & {\B 61.5}  & 59.0 & {\B 64.6} & 65.5 & {\B 67.1} & 63.4 & {\B 68.8} & 66.2 & {\B 69.0} \\
\texttt{is\_IS}   & 11.6  & {\B 32.1}  & {\B 47.2}  & 31.7  & 47.6 & {\B 60.9} & 63.4 & {\B 65.9} & 54.6 & {\B 68.5} & 63.4 & {\B 69.6} \\
\texttt{it\_IT}   & 25.3  & {\B 52.5}  & 59.5  & 59.5       & 57.2 & {\B 63.0} & 64.6 & {\B 65.5} & 60.2 & {\B 67.6} & 65.7 & {\B 67.3} \\
\texttt{ja\_JP}   & {\B 26.8}  & 23.3  & {\B 46.6}  & 29.3  & 51.0 & {\B 55.6} & 57.3 & {\B 58.8} & 60.5 & {\B 65.8} & 58.7 & {\B 67.0} \\
\texttt{jv\_ID}   & 10.7  & {\B 22.9}  & 45.8  & {\B 46.2}  & 42.5 & {\B 58.9} & 62.1 & {\B 63.9} & 48.5 & {\B 66.5} & 62.6 & {\B 68.5} \\
\texttt{ka\_GE}   & 9.7   & {\B 17.9}  & {\B 39.9}  & 22.1  & 45.4 & {\B 52.9} & 54.8 & {\B 57.1} & 54.5 & {\B 63.8} & 56.2 & {\B 66.8} \\
\texttt{km\_KH}   & 11.4  & {\B 18.0}  & {\B 44.8}  & 23.6  & 39.2 & {\B 51.8} & 51.7 & {\B 55.7} & 54.7 & {\B 63.8} & 54.3 & {\B 67.0} \\
\texttt{kn\_IN}   & 8.8   & {\B 20.2}  & {\B 41.9}  & 25.4  & 47.4 & {\B 58.6} & 55.8 & {\B 61.7} & 52.1 & {\B 63.8} & 56.6 & {\B 65.8} \\
\texttt{ko\_KR}   & 11.0  & {\B 16.3}  & {\B 49.8}  & 24.8  & 54.1 & {\B 61.5} & 65.6 & {\B 65.8} & 60.2 & {\B 68.7} & 66.4 & {\B 70.3} \\
\texttt{lv\_LV}   & 11.6  & {\B 40.3}  & {\B 51.9}  & 33.7  & 52.4 & {\B 61.2} & 63.0 & {\B 64.6} & 59.0 & {\B 69.6} & 64.1 & {\B 70.4} \\
\texttt{ml\_IN}   & 10.1  & {\B 19.4}  & {\B 41.2}  & 25.8  & 47.9 & {\B 55.3} & 55.0 & {\B 58.5} & 59.4 & {\B 68.2} & 55.6 & {\B 69.2} \\
\texttt{mn\_MN}   & 7.4   & {\B 13.4}  & {\B 38.9}  & 22.2  & 46.9 & {\B 57.0} & 60.2 & {\B 62.7} & 53.8 & {\B 66.1} & 61.5 & {\B 68.7} \\
\texttt{ms\_MY}   & 21.7  & {\B 45.0}  & 54.8  & {\B 59.9}  & 57.1 & {\B 65.7} & 67.7 & {\B 68.0} & 60.6 & {\B 69.3} & 68.4 & {\B 68.9} \\
\texttt{my\_MM}   & 10.7  & {\B 13.8}  & {\B 48.7}  & 23.1  & 51.5 & {\B 59.8} & 61.9 & {\B 66.1} & 59.3 & {\B 68.8} & 64.3 & {\B 72.6} \\
\texttt{nb\_NO}   & 26.9  & {\B 50.6}  & {\B 60.7}  & 56.3  & 60.7 & {\B 68.0} & 68.8 & {\B 70.2} & 65.0 & {\B 70.5} & 69.9 & {\B 70.7} \\
\texttt{nl\_NL}   & 28.3  & {\B 55.2}  & 60.1  & {\B 63.3}  & 60.2 & {\B 66.5} & 67.4 & {\B 67.5} & 64.7 & {\B 68.4} & 68.3 & {\B 70.0} \\
\texttt{pl\_PL}   & 19.0  & {\B 47.1}  & {\B 50.7}  & 46.0  & 56.2 & {\B 61.8} & 62.0 & {\B 63.3} & 59.7 & {\B 65.9} & 62.5 & {\B 66.5} \\
\texttt{pt\_PT}   & 28.1  & {\B 52.0}  & {\B 60.8}  & 50.6  & 61.5 & {\B 65.9} & 66.8 & {\B 67.6} & 63.6 & {\B 68.7} & 67.5 & {\B 68.2} \\
\texttt{ro\_RO}   & 22.8  & {\B 45.7}  & {\B 57.4}  & 52.7  & 55.8 & {\B 64.5} & 65.7 & {\B 67.1} & 60.2 & {\B 68.5} & 65.9 & {\B 69.6} \\
\texttt{ru\_RU}   & 19.0  & {\B 26.1}  & {\B 49.0}  & 26.1  & 56.9 & {\B 61.6} & 63.5 & {\B 63.8} & 63.5 & {\B 68.8} & 64.0 & {\B 69.5} \\
\texttt{sl\_SL}   & 15.8  & {\B 43.7}  & {\B 52.8}  & 47.8  & 53.2 & {\B 63.5} & 64.5 & {\B 64.8} & 57.7 & {\B 68.0} & 64.5 & {\B 68.8} \\
\texttt{sq\_AL}   & 15.3  & {\B 42.1}  & {\B 48.0}  & 39.9  & 48.8 & {\B 61.1} & 61.2 & {\B 63.5} & 54.2 & {\B 68.9} & 61.3 & {\B 68.5} \\
\texttt{sv\_SE}   & 26.0  & {\B 54.4}  & {\B 61.8}  & 53.0  & 62.6 & {\B 70.1} & 70.6 & {\B 71.1} & 65.9 & {\B 72.0} & 71.2 & {\B 71.5} \\
\texttt{sw\_KE}   & 9.6   & {\B 15.6}  & {\B 44.0}  & 41.9  & 44.2 & {\B 58.7} & 58.2 & {\B 59.6} & 48.0 & {\B 66.3} & 58.6 & {\B 66.8} \\
\texttt{ta\_IN}   & 10.9  & {\B 19.9}  & {\B 41.1}  & 24.3  & 48.2 & {\B 55.5} & 56.4 & {\B 58.3} & 56.6 & {\B 64.9} & 58.0 & {\B 66.0} \\
\texttt{te\_IN}   & 7.8   & {\B 21.6}  & {\B 46.4}  & 25.1  & 43.6 & {\B 60.0} & 55.4 & {\B 62.7} & 51.4 & {\B 65.0} & 55.1 & {\B 67.5} \\
\texttt{th\_TH}   & 21.8  & {\B 31.3}  & {\B 55.0}  & 26.8  & 47.4 & {\B 62.1} & 62.2 & {\B 66.9} & 63.2 & {\B 72.0} & 64.6 & {\B 74.2} \\
\texttt{tl\_PH}   & 18.9  & {\B 42.0}  & 56.9  & {\B 58.7}  & 53.2 & {\B 62.4} & 65.7 & {\B 66.1} & 56.7 & {\B 66.5} & 66.5 & {\B 68.5} \\
\texttt{tr\_TR}   & 14.4  & {\B 35.2}  & {\B 48.4}  & 38.5  & 51.6 & {\B 64.9} & 65.5 & {\B 66.2} & 58.5 & {\B 69.4} & 65.5 & {\B 69.4} \\
\texttt{ur\_PK}   & 9.7   & {\B 22.7}  & {\B 49.2}  & 22.8  & 50.5 & {\B 59.5} & 61.5 & {\B 61.9} & 54.1 & {\B 63.3} & 62.6 & {\B 65.7} \\
\texttt{vi\_VN}   & 15.1  & {\B 35.1}  & {\B 55.9}  & 36.4  & 49.8 & {\B 57.5} & 61.0 & {\B 62.3} & 55.5 & {\B 67.0} & 62.1 & {\B 68.2} \\
\texttt{zh\_CN}   & {\B 22.1}  & 17.3  & {\B 31.7}  & 24.1  & 45.6 & {\B 54.1} & 53.0 & {\B 57.9} & 60.8 & {\B 65.9} & 54.9 & {\B 66.6} \\
\texttt{zh\_TW}   & {\B 21.2}  & 16.5  & {\B 32.4}  & 24.2  & 45.2 & {\B 51.8} & 52.0 & {\B 54.5} & 58.2 & {\B 62.2} & 53.8 & {\B 63.9} \\
\midrule
\textbf{Average}  & 17.7  & {\B 33.5}  & {\B 50.2}  & 38.3  & 51.8 & {\B 61.2} & 62.5 & {\B 64.2} & 58.2 & {\B 67.5} & 63.3 & {\B 68.7} \\
\bottomrule
\end{tabular}

\caption{*T5 parsers Exact Match on individual languages in the Zero-Shot, \taf{} and Gold settings.}
\label{table:full-results}
\end{table*}

\end{document}